\documentclass{article}

\usepackage{PRIMEarxiv}

\usepackage[utf8]{inputenc} 
\usepackage[T1]{fontenc}    
\usepackage{hyperref}       
\usepackage{url}            
\usepackage{booktabs}       
\usepackage{amsfonts}       
\usepackage{nicefrac}       
\usepackage{microtype}      
\usepackage{lipsum}
\usepackage{fancyhdr}       
\usepackage{graphicx}       
\graphicspath{{media/}}     
\usepackage{epstopdf}
\usepackage{multirow}
\usepackage{tabularx}
\usepackage{amsmath, amssymb}
\usepackage{makecell}
\usepackage{adjustbox}
\usepackage{xcolor}
\usepackage{tablefootnote}

\title{An Overview of Language Models: \\Recent Developments and Outlook
}

\usepackage{authblk}
\author[1]{Chengwei Wei}
\author[1]{Yun-Cheng Wang}
\author[2]{Bin Wang}
\author[1]{C.-C. Jay Kuo}
\affil[1]{University of Southern California, Los Angeles, California, USA}
\affil[2]{National University of Singapore, Singapore}

{
    \makeatletter
    \renewcommand\AB@affilsepx{: \protect\Affilfont}
    \makeatother

    \makeatletter
    \renewcommand\AB@affilsepx{, \protect\Affilfont}
    \makeatother

    \affil[ ]{\texttt{chengwei@usc.edu}}
    
}

\begin{document}
\maketitle

\begin{abstract}

Language modeling studies the probability distributions over strings of
texts. It is one of the most fundamental tasks in natural language
processing (NLP). It has been widely used in text generation, speech
recognition, machine translation, etc. Conventional language models (CLMs) aim to predict
the probability of linguistic sequences in a causal manner, while
pre-trained language models (PLMs) cover broader concepts and can be
used in both causal sequential modeling and fine-tuning for downstream
applications.  PLMs have their own training paradigms (usually
self-supervised) and serve as foundation models in modern NLP systems.
This overview paper provides an introduction to both CLMs and PLMs from
five aspects, i.e., linguistic units, architectures, training methods,
evaluation methods, and applications. Furthermore, we discuss the
relationship between CLMs and PLMs and shed light on the future
directions of language modeling in the pre-trained era. 

\end{abstract}


\section{Introduction}\label{LMsection:introduction}

Language modeling studies the probability distributions over a sequence
of linguistic units, such as words. It is one of
the most fundamental tasks and long-standing research topics in natural
language processing (NLP).  The developed language models (LMs) find
applications in many computational linguistic problems such as text
generation, machine translation, speech recognition, natural language
generation, question-and-answer systems, etc. 

There are two major approaches to language modeling: 1) the statistical
approach based on a relatively small corpus set, and 2) the data-driven
approach based on a significantly larger corpus set.  Conventional
language models (CLMs) predict the probability of linguistic sequences
in a causal manner. They can be learned by both language modeling
approaches. The data-driven approach has become mainstream nowadays.  It
exploits a large number of corpora to train neural-network models,
leading to pre-trained language models (PLMs). PLMs are then fine-tuned
with task-specific datasets and objectives for downstream applications.
In this overview paper, we define CLMs as language
models that predict the probability of linguistic sequences in a causal
manner. In contrast, PLMs refer to language models pre-trained on a
broad range of linguistic tasks and objectives. It is important to note
that the two concepts are not exclusive. One LM can fall into both
categories. For example, GPT models \cite{radford2018improving} can
predict the probability of linguistic sequences in a causal manner. They
are also pre-trained with various downstream tasks. We provide an
overview of CLMs and PLMs and study them from five perspectives: 1)
linguistic units, 2) architectures, 3) training methods, 4) evaluation
methods, and 5) applications.  In the end, we point out several future
research directions. 

The goal of CLMs is to model the probability distributions over
sequences of linguistic units:
\begin{equation} \label{eq:lm}
P(u_1, u_2, \cdots ,u_{t}),
\end{equation}
where $u_i$ can be either a character, a word, a phrase, or other linguistic
units. CLMs attempt to predict the next linguistic unit in a
text sequence given its preceding contexts:
\begin{equation} \label{eq:lm3}
P(u_t|u_{<t})
\end{equation}
CLMs are also called auto-regressive language models since the units are
predicted in a causal way. Estimating the probability of a text sequence
as shown in Eq.~(\ref{eq:lm}) directly encounters the data sparsity
problem.  CLMs often estimate the joint probability of the text sequence
by decomposing a text sequence into smaller units.  For example, CLMs
leverage the chain rule and the conditional probability to estimate the
joint probability in the form of
\begin{equation} \label{eq:lm2}
P(u_1, u_2, \cdots ,u_t) = P(u_1) P(u_2|u_1) P(u_3|u_1,u_2) \cdots 
P(u_{t}|u_1,...u_{t-1}).
\end{equation}

Before the pre-training era, CLMs are often trained
from scratch with a training corpus and, then, predict the probability
of text sequences with respective applications.  Representative models
include N-grams LMs~\cite{brown1992class, federico1996bayesian,
niesler1996variable}, exponential LMs ~\cite{della1992adaptive,
berger1996maximum, rosenfeld1996maximum} and earlier neural
LMs~\cite{bengio2000neural, mikolov2010recurrent}. CLMs give a high
probability to natural text sequences occurring frequently in the real
world. As a result, they play a fundamental role in text generation,
speech recognition~\cite{jelinek1975design, jelinek1976continuous,
bahl1983maximum}, and machine translation~\cite{brown1990statistical,
och2001efficient, yamada2002decoder} until the emergence of PLMs.
Nowadays, high-performance PLMs serve as the backbone of many NLP
systems. They are not limited to the causal predictive functionality of
CLMs and provide more different types of LMs. 

The differences between CLMs before the pre-training
era and PLMs can be summarized below. 
\begin{itemize}
\item {\bf Training Methodology.} With the development of deep learning,
PLMs with neural network structures are pre-trained by collections of
massive unlabeled corpora to learn generic knowledge which is then
transferred to downstream tasks by task-specific fine-tuning.  \item
{\bf Causality Constraint.} PLMs do not necessarily follow CLMs in
predicting linguistic units as shown in Eq.~(\ref{eq:lm3}).  For
example, bidirectional LMs~\cite{devlin2018bert, liu2019roberta} use
both preceding and succeeding contexts to predict the missing linguistic
units via probability estimation:
\begin{equation} \label{eq:blm}
P(u_t|u_{<t}, u_{>t}).
\end{equation}
Bidirectional LMs do not follow the causality constraint and the
chain rule in Eq.~(\ref{eq:lm2}), to access the probability of a text
sequence, which makes it inherently different from CLMs. 
\item {\bf Token Representation.} Apart from the differences in the
training paradigm and probability modeling, PLMs adopt a different
representation for basic units called tokens. PLMs represent tokens
by embedding them in a high-dimensional continuous space such as
word embeddings~\cite{peters-etal-2018-deep, wei2023synwmd} and sentence
embeddings~\cite{wang2020sbert,gao2021simcse, wang2022relational}.  The new
representations offer a flexible and powerful tool that enables PLMs to handle
a wide range of tasks.
\end{itemize}

This overview paper serves two objectives. On one hand, instead of only focusing on recently developed PLMs ~\cite{qiu2020pre, han2021pre,
liu2023pre}, we aim to
provide a comprehensive overview of the basic concepts of LMs, the transition
from CLMs to PLMs, LM's recent developments and applications to
beginners in the field. On the other hand, we would like to shed light
on future research directions and offer our outlook to experienced
engineers and researchers in the NLP field. For example, we cover large
LMs (LLMs) in the survey as there are growing interests in LLMs due to
the new services provided by ChatGPT.  Furthermore, we include efficient
LMs as an emerging topic since there are increasing concerns about large
model sizes and high training costs of LLMs.

The rest of the paper is organized as below.  We introduce several types of
LMs that go beyond CLMs in Sec.~\ref{LMsec:types}, and provide an
overview of common ways to decompose text sequences into smaller
linguistic units in Sec.~\ref{LMsec:LU}.  Sec.~\ref{LMsec:structures}
introduces different model architectures.  We discuss the training
procedures of LMs in Sec.~\ref{LMsec:PLM}.  Common evaluation methods
including, both intrinsic and extrinsic ones, are introduced in
Sec.~\ref{LMsec:Evaluation}.  The application of LMs to text generation
is discussed in Sec.~\ref{LMsec:Application}.  We comment on the
redundancy problem of LMs and analyze techniques for efficient LMs in
Sec.~\ref{LMsec:Efficient_LMs}.  Promising future research directions
are pointed out in Sec.~\ref{LMsec:future_directions}.  Concluding
remarks are given in Sec. \ref{LMsec:conclusion}

\section{Types of Language Models}\label{LMsec:types}

CLMs commonly refer to auto-regressive models that predict the next
linguistic units given the preceding context as shown in
Eq.~(\ref{eq:lm3}). LMs can access the probability of a text sequence
using the chain rule. The goal of CLMs is to decode the probability of
text sequences in a causal manner. In this section, we introduce more
LMs that go beyond CLMs. 

\subsection{Structural LM} 

\begin{figure}[t]
\centering
\includegraphics[width=0.55\textwidth]{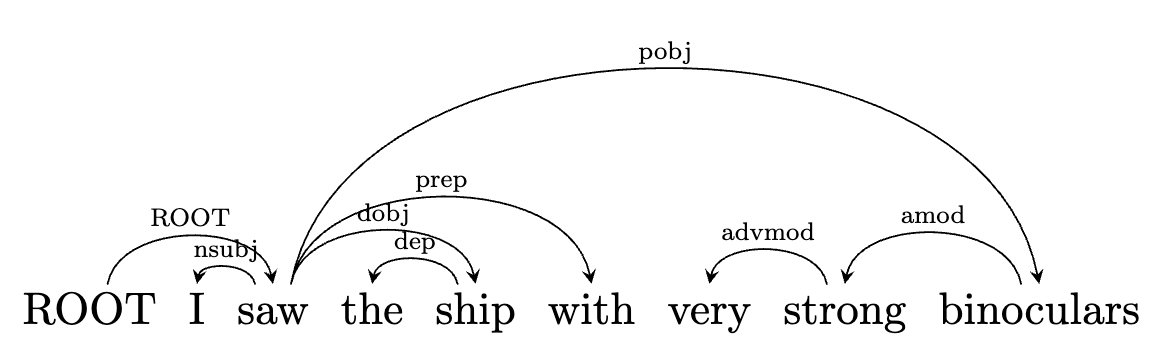}
\caption{The example of a dependency parse tree example~
\cite{mirowski2015dependency}.}\label{fig:parse tree}
\end{figure}

Instead of predicting linguistic units in a sequential or reversed
sequential order, structural LMs~\cite{chelba1998exploiting,
chelba2000structured, gubbins2013dependency, mirowski2015dependency,
wei2022task} predict linguistic units based on pre-defined linguistic
structures such as dependency or constituent parse trees. Structural LMs
utilize the linguistic structure to bring linguistically relevant
context closer to the linguistic unit to be predicted. For example,
given a parse tree structure, a structural LM can define the ancestor
context $A(u_t)$ of $u_t$ as the sequence from the root node to the
parent of $u_t$. For example, the ancestor sequence of word `strong' is
\{`binoculars', `saw', ROOT\} in Fig.~\ref{fig:parse tree}.  Then, the
structural LM uses the ancestor context in the tree to predict the next
linguistic unit as
\begin{equation}\label{eq:slm}
P(u_t|A(u_t)), 
\end{equation} 
where $A(u_t)$ is the ancestor context of linguistic unit $u_t$. 
Similar to CLMs, structural LMs are designed to model the probability of
text sequences. Differently, structural LMs decode the sequence
probability in the order of their synthetic structures. It has been
successfully applied to sentence completion~\cite{gubbins2013dependency,
mirowski2015dependency} and speech recognition~\cite{chelba1998exploiting, 
chelba2000structured}. 

\subsection{Bidirectional LM}

Instead of using the causal contexts to make predictions, bidirectional LMs
utilize contexts from both directions as shown in Eq.~(\ref{eq:blm}).
The masked LM is one representative bidirectional LM. It masks out
linguistic units in a text sequence and, then, encodes their preceding
and succeeding contexts to predict the masked linguistic units.
Formally, the prediction can be defined as the estimation of the following
conditional probability
\begin{equation} \label{eq:mlm}
P(u_m|\bar{S}),
\end{equation}
where $u_m$ is the masked linguistic unit and $\bar{S}$ is the corrupted
text sequence by replacing a certain number of linguistic units with
\texttt{[MASK]} symbols. The goal of bidirectional LMs is to learn the
inner dependency between linguistic units in an unsupervised manner. The
trained model can inherit semantics meanings from large-scale unlabeled
corpora. Different from CLMs that aim to model the generation
probability of text sequences, pre-trained bidirectional LMs are used as
the backbone that transfers the learned knowledge through further
fine-tuning in various downstream applications. 

\begin{figure}[t]
\centering
\includegraphics[width=0.4\textwidth]{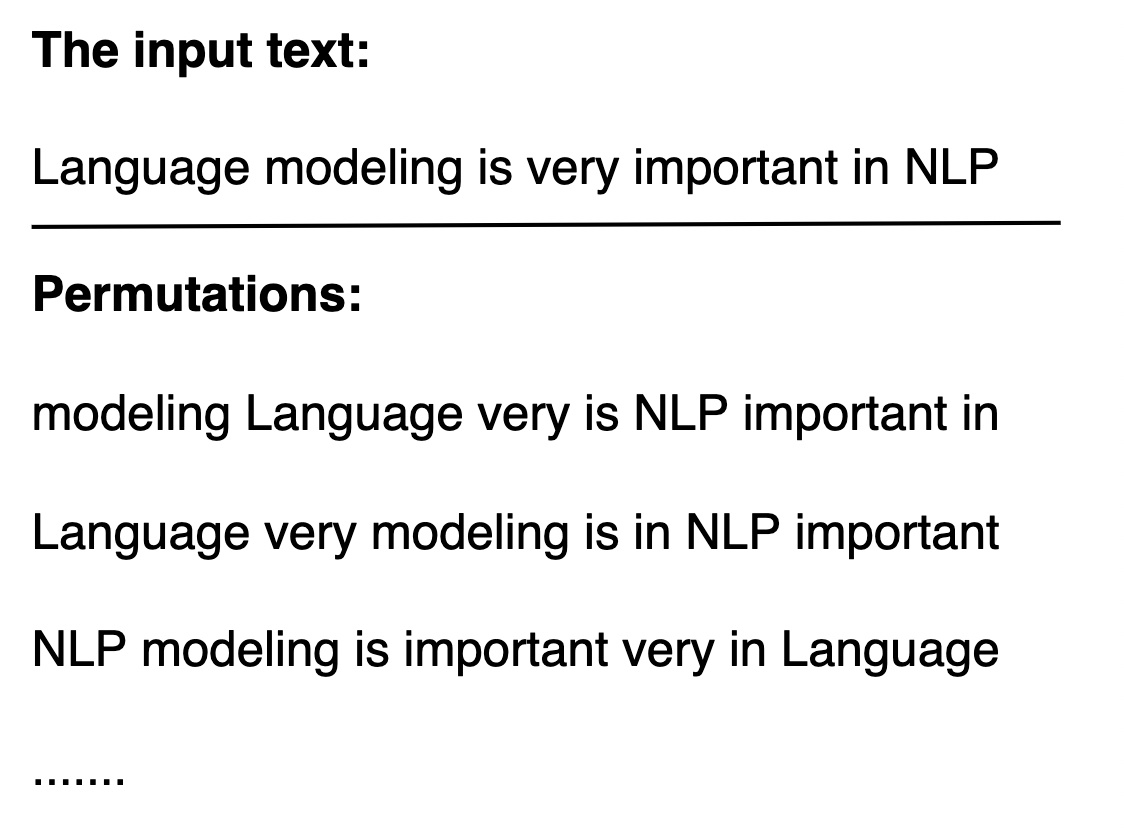}
\caption{The use of different permutations in a natural sentence.}
\label{fig:permutation}
\end{figure}

\subsection{Permutation LM}

CLMs and masked LMs have their own advantages and disadvantages. A
masked LM needs to create artificial tokens such as \texttt{[mask]},
which never occur in downstream tasks while CLMs only condition on
preceding context. The permutation LM~\cite{yang2019xlnet} is a recently
proposed LM that takes advantage of CLMs and masked LMs. Given an input
sequence of linguistic units, permutation LMs randomize the order of
input linguistic units and construct different permutations of the input
sequence.  Fig.~\ref{fig:permutation} shows an example of different
permutations given an input text sequence.  Let $\mathbb{Z}$ be the set
of all possible permutations. Permutation LMs predict the next
linguistic unit, $u_t$, in one permutation, $Z$, of the sequence based on
\begin{equation} \label{eq:permutation lm}
P(u_t|u_{<t}^{Z}), Z \in\mathbb{Z}.
\end{equation}

\section{Linguistic Units} \label{LMsec:LU}

To estimate the probability of text sequences, LMs partition text
sequences into small linguistic units such as characters, words,
phrases, or sentences. This process is called tokenization.
The resulting linguistic units are called tokens.
Different languages and models may have different appropriate
tokenization methods.  Here, we focus on English and use it as an
example. In this section, we examine typical tokenization methods used
in language modeling according to unit sizes. 

\subsection{Characters}

LMs can model text sequences probability based on
characters~\cite{sutskever2011generating, hwang2017character,
kim2016character, al2019character, xue2022byt5}. As compared with other
linguistics units, using characters has a much smaller vocabulary size,
leading to a smaller discrete space and model size. On the other hand,
it is challenging to predict the next character. Usually, it requires a
long historical context. This makes the performance of character-level
LMs poorer than that of word-level LMs.  In addition, the input and
output lengths have to be longer to model the character distribution
accurately. This results in higher computational costs, especially for
auto-regressive decoding. Several LM methods use the combination of
words and characters to alleviate the issue~\cite{kang2011mandarin,
miyamoto2016gated, verwimp2017character}. 

\subsection{Words and Subwords}

The most natural tokenization for English is to decompose a text
sequence into words by white spaces. Many LMs apply word tokenization.
However, there are several issues of naive word tokenization. The first
one is the Out-Of-Vocabulary (OOV) problem.  Because an LM has a
pre-defined vocabulary size that cannot be arbitrarily large. Less
frequent words and words with character-level errors may not be stored
in the pre-defined vocabulary. Thus, they cannot be retrieved from the
dictionary. Although one can extend the vocabulary size to alleviate
this problem, it will increase the model size and still cannot handle
all possible words. 

LMs beyond the word level still have the OOV problem while a single
character is not semantically meaningful by themselves.  Recently,
researchers are in favor of decomposing words into subwords if they do
not appear in the dictionary. This offers a flexible and effective
solution to the OOV problem~\cite{mikolov2012subword,
sennrich2016neural}. Several subword segmentation algorithms are
developed to boost the performance of LMs. They strike a balance between
the good performance of word-level models and the flexibility of
character-level models. Two subword segmentation approaches,
statistics-based and linguistics-based, are presented below. 

\subsubsection{Statistics-based Subword Tokenizers}

The statistics-based subword tokenizers generate subword vocabulary
purely based on the corpus. The associated methods are derived from a
compression point of view. They work by replacing the commonly appeared
character sequences with a new symbol (word) that does not exist in the
current vocabulary.  Then, fewer bytes are needed for information
transmission. 
  
\begin{figure}[t]
\centering
\includegraphics[width=0.8\textwidth]{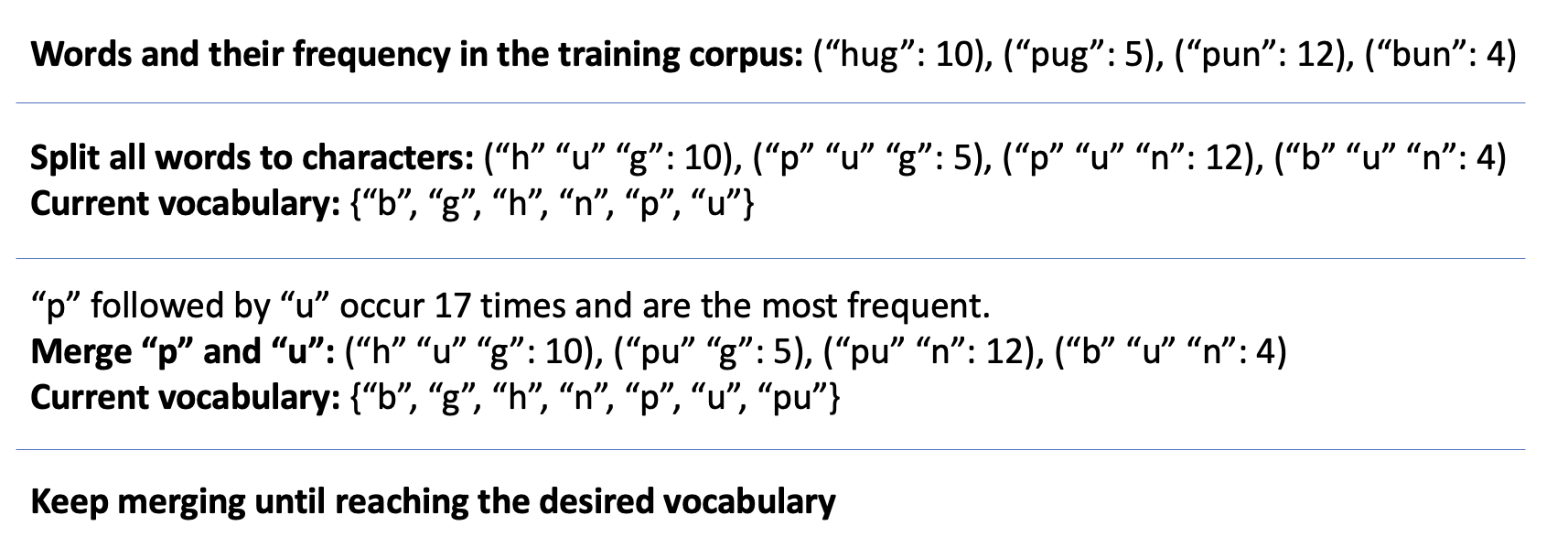}
\caption{Illustration of the BPE merge operation conducted on the
dictionary \{``hug", ``pug", ``pun", ``bun"\}. The vocabulary is
initialized with all characters.  Then, a new subword is created by
merging the most frequent pair.}\label{fig:subword_BPE}
\end{figure}

{\bf Byte Pair Encoding (BPE).} BPE~\cite{gage1994new} is a simple data
compression technique that replaces the most common pair of bytes in a
sequence by a single unused byte recursively. It was adopted by
~\cite{sennrich2016neural} to solve the word segmentation problem.  That
is, frequent characters or character sequences are merged to generate
subwords. BPE is also used by several advanced PLMs such as
GPT-2~\cite{radford2019language} and RoBERTa~\cite{liu2019roberta} with
the following algorithm, called the BPE merge operation.
\begin{enumerate} 
\item Prepare a training corpus and define the size of the subword vocabulary. 
\item Split all words into characters. 
\item Generate a new subword by merging a pair of characters or subwords
with the highest frequency. 
\item Repeat step 3 until the desired vocabulary size is reached.
\end{enumerate}
An illustration of the BPE merge operation conducted on a small dictionary 
is given in Fig.~\ref{fig:subword_BPE}.

{\bf WordPiece.}~\cite{schuster2012japanese} WordPiece is another
data-driven subword algorithm. The difference between WordPiece and BPE
is that WordPiece merges the pair of $A$ and $B$ if they have the
highest score $P(AB)/P(A)P(B)$ (rather than the highest frequency
$P(AB)$) at each iterative step. For example, WordPiece merges the pair
of ``u'' and ``g'' in Fig.~\ref{fig:subword_BPE} only if they have the
highest value, $P('ug')/P('u')P('g')$, as compared with other pairs.
WordPiece is used as the tokenization method in
BERT~\cite{devlin2018bert}, DistilBERT~\cite{sanh2019distilbert}, and
Electra~\cite{clark2020electra}. 

There are other statistics-based subword tokenizers such as
{\bf{Unigram}}~\cite{kudo2018subword}.
SentencePiece~\footnote{\url{https://github.com/google/sentencepiece}},
Huggingface
tokenizers~\footnote{\url{https://github.com/huggingface/tokenizers}},
and OpenNMT~\footnote{\url{https://github.com/OpenNMT/Tokenizer}} are
popular tokenizers. Their implementation contains the statistics-based
subword tokenization. Different subword tokenizers and their
performance comparison are studied in~\cite{bostrom2020byte}.

\subsubsection{Linguistics-based Subword Tokenizers}

Linguistics-based subword tokenizers exploit the linguistic knowledge
and decompose words into smaller grammatical units, such as morphemes or
syllables. Such subword tokenizers are widely used in machine
translation and speech recognition among different
languages~\cite{kiecza1999data, creutz2005unsupervised, creutz2007morph,
rotovnik2007large, sarikaya2007joint, sak2010morphology,
ablimit2010uyghur}. For example, in machine translation, words formed by
compounding, affixation, or inflection, can be conveniently translated
by translating the morphemes, respectively. However, linguistics-based
subword tokenizers are not as popular as statistics-based ones due to
the complexity and the rule-based nature of language decomposition. 

\subsection{Phrases}

The semantic meaning of a single word can be ambiguous because of
various contexts and set collocations. Since the linguistic dictionary
does not go beyond the word-level, the inter-word dependency is ignored.
Phrase-level LMs replace common and cohesive word sequences by
phrases~\cite{suhm1994towards, ries1996class, saon2001data,
levit2014word}. Phrase-level LMs are suitable for some applications. For
example, it is observed in \cite{saon2001data} that short words with
fewer syllables in automatic speech recognition (ASR) are more
frequently misrecognized than longer ones. Since phrases provide longer
phone sequences than their constituents, they are more robust to
recognition errors for ASR. 

\subsection{Sentences}

Auto-regressive LMs with smaller linguistic units
(e.g., characters, words, subwords, and phrases) rely on conditional
probabilities to estimate the probability of text sequences as given in
Eq.~(\ref{eq:lm2}).  Sentence-level LMs~\cite{rosenfeld1997whole,
chen1999efficient, rosenfeld2001whole, ippolito2020toward, lee2020slm}
avoid the use of the chain rule. They generate sentence features and,
then, model the sentence probability directly. This is because modeling
the sentence probability directly is more convenient than that in
Eq.~(\ref{eq:lm2}) in encoding the sentence-level information.  It is
also easier to encode the inter-sentence information such as the effects
of preceding utterances in a dialog flow. 

\section{Architecture of Language Models}\label{LMsec:structures}

In this section, we conduct a survey on several common architectures to
model the probability distributions of text sequences.  They are N-gram,
maximum entropy, and neural network models. While
there are other LM architectures, like Gaussian mixture LMs
\cite{afify2007gaussian} and Hidden Markov LMs \cite{kuhn1994ergodic},
we focus on the above-mentioned architectures due to their popularity in
the research community. Furthermore, LMs can operate at various levels
of linguistic units. For generality and consistency with most recent
literature, we use the term `token' to refer to all linguistic units
leveraged by different LMs for the rest of this paper.

\subsection{N-gram Models}

An N-gram consists of N consecutive tokens from a
text sequence. N-gram LMs~\cite{brown1992class, federico1996bayesian,
niesler1996variable} assume that the probability of a
token depends only on its preceding N-1
tokens and it is independent of other contexts.
This is known as the Markov assumption.  Thus, instead of using all
historical contexts, N-gram LMs only use the previous N-1
tokens to predict the current one; namely,
\begin{equation}
\label{eq:ngram lm} P(u_t|u_{<t}) = P(u_t|u_{t-N+1:t-1}).
\end{equation}
N-gram LMs calculate the conditional probability by counting
the occurrence time of N-grams given a training corpus as
\begin{equation} \label{eq:ngram lm2}
P(u_t|u_{t-N+1:t-1})=\frac{C(u_{t-N+1:t})}{C(u_{t-N+1:t-1})}.
\end{equation}

N-gram LMs simplify the token probability
calculation based on previous N-1 tokens, but they
encounter two sparsity issues. First, if an N-gram, $(u_{t-N+1:t})$,
never occurs in the training corpus, the probability for the next
tokens being $u_t$ is zero. Second, if the
(N-1)-gram, $(u_{t-N+1:t-1})$, in the denominator never occurs, we
cannot calculate the probability of any tokens.
These sparsity issues can be alleviated by smoothing techniques.  A
simple smoothing method~\cite{lidstone1920note, johnson1932probability},
called additive smoothing, is to add a small value to the count for
every N-gram so as to avoid zero in the numerator and the denominator in
Eq.~(\ref{eq:ngram lm2}). However, this simple smoothing is still
deficient because it assigns the same probability for N-grams that never
occur in the training corpus. 

There are more advanced smoothing techniques such as back-off and
interpolation~\cite{jelinek1980interpolated, katz1987estimation,
church1991comparison, kneser1995improved, chen1999empirical} that
achieve better probability estimation. In back-off, lower-order N-grams
are used for probability estimation if higher-order N-grams do not
occur.  For example, if $C(u_{t-3:t-1})=0$, we back off to compute
$P(u_t|u_{t-2:t-1})$. In interpolation, different N-grams are considered
for conditional probability computation. Mathematically, the N-gram
probability is estimated by
\begin{multline} \label{eq:interpolation}
P(u_t|u_{t-N+1:t-1})=\lambda_N P(u_t|u_{t-N+1:t-1})+\lambda_{N-1} P(u_t|u_{t-N:t-1}) \\ 
+\lambda_{N-2} P(u_t|u_{t-N-1:t-1})+...+\lambda_{1} P(u_t),
\end{multline}
where $\lambda_i$ is the weight for each n-gram and $\sum_{i=1}^{N} \lambda_{i}=1$.

\subsection{Maximum Entropy Models}

Maximum Entropy models (also called the exponential models)~\cite{della1992adaptive, 
berger1996maximum, rosenfeld1996maximum} estimate the probability of text sequences 
using feature functions in the form of
\begin{equation} \label{eq:max_entropy lm}
P(u|h)=\frac{exp(a^T f(u,u_{<t}))}{\sum_{u'}^{}exp(a^T f(u',u'_{<t}))},
\end{equation}
where $f(u, u_{<t})$ is the feature function that generates the feature
of token $u$ and its historical context $u_{<t}$,
$\sum_{w'}^{}exp(a^T f(u',u'_{<t}))$ is a normalization factor, and $a$
is a parameter vector derived by the Generalized Iterative Scaling
algorithm~\cite{darroch1972generalized}. The features are usually
generated from the N-grams. 

\subsection{Feed-forward Neural Network (FNN) Models}

The discrete nature of the N-gram model is its performance bottleneck
even with advanced smoothing techniques. Neural LMs embrace the
continuous embedding space (distributed representation) to
overcome the data sparsity problem.  Feed-forward Neural Network (FNN)
LMs~\cite{bengio2000neural, schwenk2005training, schwenk2007continuous,
arisoy2012deep} is one of the earlier neural network models. 

\begin{figure}[t]
\centering
\includegraphics[width=0.9\linewidth]{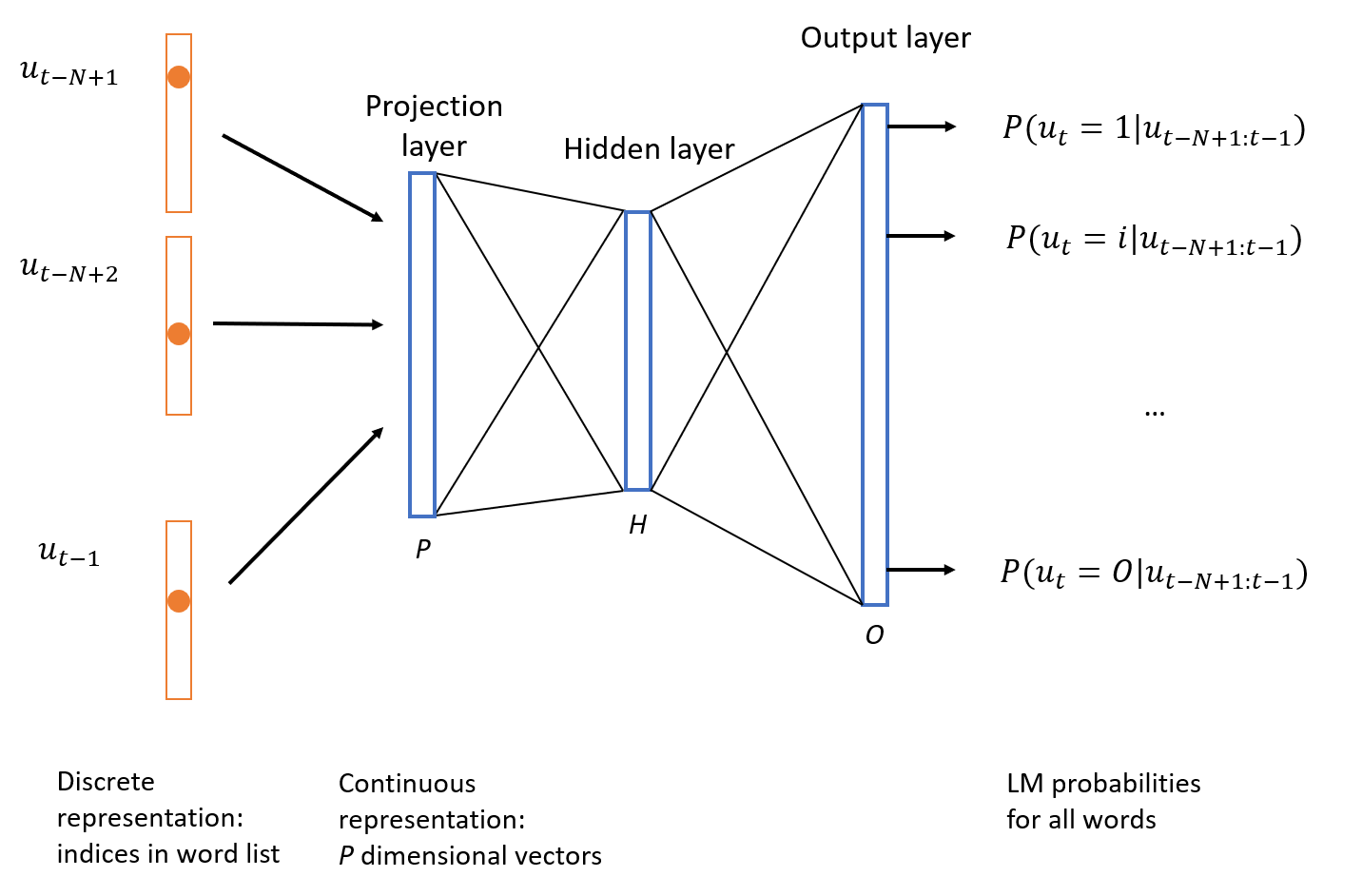}
\caption{ The structure
of FFN LMs, where $u_{t-N+1},...,u_{t-1}$ denotes the preceding contexts of $u_{t}$ in a fixed-window, and $P$, $H$, and $O$ are the
dimensions of the projection, the hidden layer, and the output layer,
respectively.}\label{fig:ffnn}
\end{figure}

An FNN LM takes historical contexts as the input, and outputs the
probability distribution of tokens. As shown in
Fig.~\ref{fig:ffnn}, each token in the preceding
context is represented as a vector through a projection layer (i.e., an
embedding matrix). These vectors of tokens are sent
to the hidden layer with $H$ hidden units followed by non-linear
activation. Then, a softmax function is used to obtain the posterior
probabilities for token candidates, $P(u_t =
i|u_{t-N-1:t-1})$, which are the probabilities of tokens
given a specific history predicted by the language model. 

An FNN LM uses a fixed window to collect fixed-length contexts. It is
essentially a neural version of N-gram LMs. The FNN LM have several
advantages over the N-gram LM by projecting tokens
into continuous space.  First, it can handle unseen N-grams by
representing each token as an N-gram with a dense
vector space. Second, it is storage-efficient since it does not need to
count and store the transition probability of conventional N-gram
models. 

\subsection{Recurrent Neural Network (RNN) Models}

It is clearly insufficient to use the historical context in a
fixed-length to predict the next token. In contrast
to the limited historical context used in the N-gram,
maximum entropy and FNN LMs, Recurrent Neural
Network (RNN) LMs~\cite{mikolov2010recurrent, mikolov2011extensions,
kombrink2011recurrent, sundermeyer2012lstm, yang2017breaking} can
exploit arbitrarily long histories to predict the next
token. 

\begin{figure}[ht]
\centering
\includegraphics[width=0.8\linewidth]{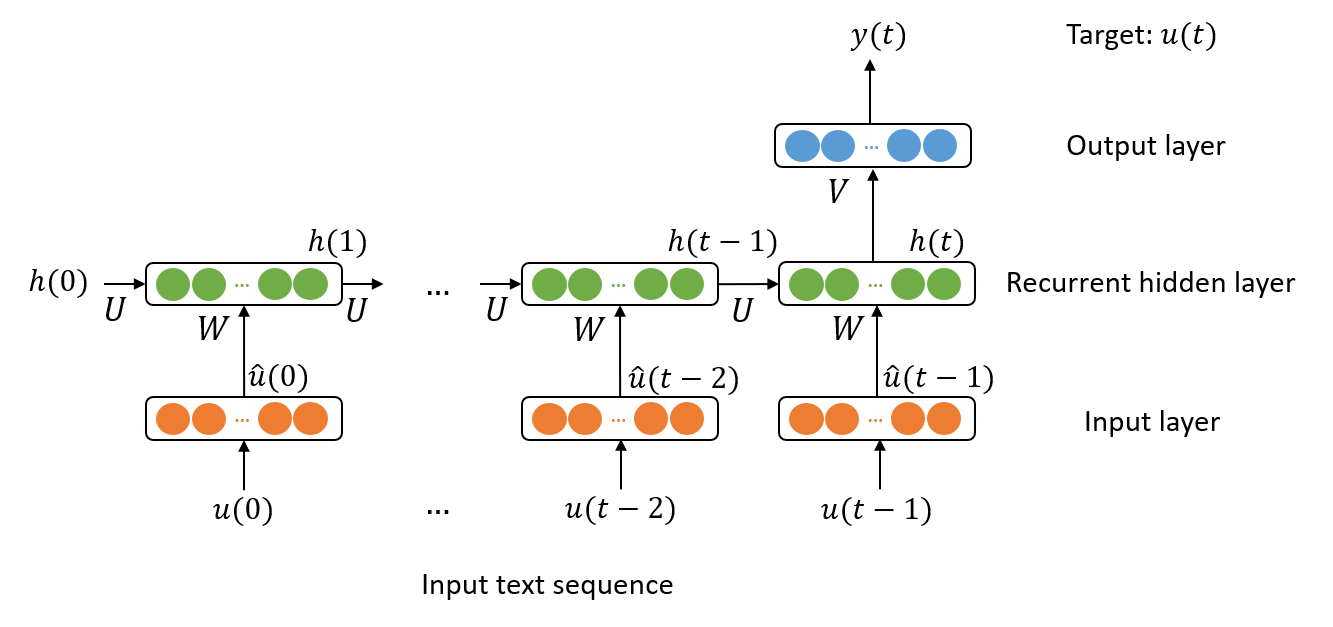}
\caption{The structure of RNN LMs.}\label{fig:rnn}
\end{figure}

The structure of a vanilla RNN LM is shown in Fig.~\ref{fig:rnn}. A
token $u(i)$ in position $i$ is first converted
into a one-hot representation $\hat{u}(i)$. Then, the recurrent hidden
state, $h(i+1)$, is computed using the previous hidden state, $h(i)$,
and the one-hot representation, $\hat{u}(i)$, of
token $u(i)$ as
\begin{equation} \label{eq:rnn}
h(i+1) = f(W\hat{u}(i)+Uh(i)),
\end{equation}
where $f(\cdot)$ is a non-linear activation function, $W$ is the weight
matrix of the connections from the input layer to the hidden layer, and
$U$ is the connection between the previous and current hidden layers,
respectively. By iteratively computing the hidden states, RNN LMs can
encode the historical context of varying length. Finally, the output
layer gives the conditional probability of tokens $y(t) = g(Vh(t))$,
where $V$ is the weight matrix connecting the hidden layer and the
output layer and $g(\cdot)$ is the softmax activation function. 

In theory, RNN LMs do not need the Markov assumption. They can use all
preceding history to predict the next token.
However, the inherent gradient vanishing problem of RNN hampers the
learning of the model~\cite{hochreiter1998vanishing}. Since the gradient
may become very small over a long distance, model weights are actually
updated by the nearby context in practice. Generally, RNN LMs cannot
learn the dependency between the current token and
its far-away historical context. Although an attention mechanism can be
introduced to RNNs to alleviate this problem \cite{bahdanau2014neural,
deng2019global}.  The inherent sequential nature of RNNs makes them less
powerful than transformer-based LMs with a self-attention mechanism. 

\begin{figure}[ht]
\centering
\includegraphics[width=0.65\linewidth]{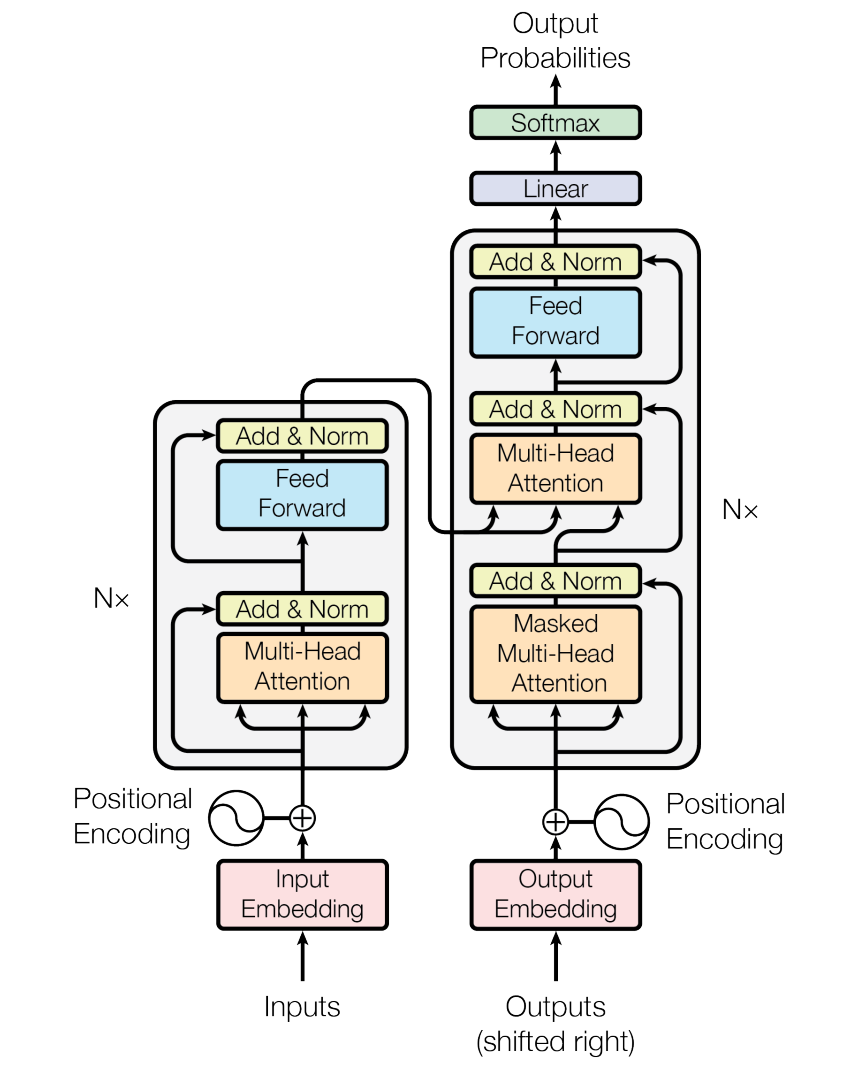}
\caption{The structure of a transformer~\cite{vaswani2017attention}.}
\label{fig:transformer}
\end{figure}

\subsection{Transformers}

The transformer architecture~\cite{vaswani2017attention} can capture
long-term dependencies and important sequence components by exploiting a
self-attention mechanism. Unlike the recurrent structure of RNNs,
a transformer is easy to parallelize in both training and inference.  Its
structure is shown in Fig.~\ref{fig:transformer}. It consists of an
encoder and a decoder.  Before being sent to the encoder, the input
textual sequence is first converted to an embedding through an embedding
layer plus positional embedding. Multi-head attention, which is an
ensemble of multiple self-attention mechanisms, enables the transformer
to capture more robust and diverse attention between tokens. The other
parts in the transformer encoder include feed-forward layers, residual
connections, and normalization layers. The difference between the
transformer encoder and decoder is that the transformer decoder has an
additional masked multi-head attention layer. The masking ensures the
decoder can only access preceding tokens of the current one, which makes
the decoder auto-regressive. 

\begin{figure*}[ht]
\centering
\includegraphics[width=1\linewidth]{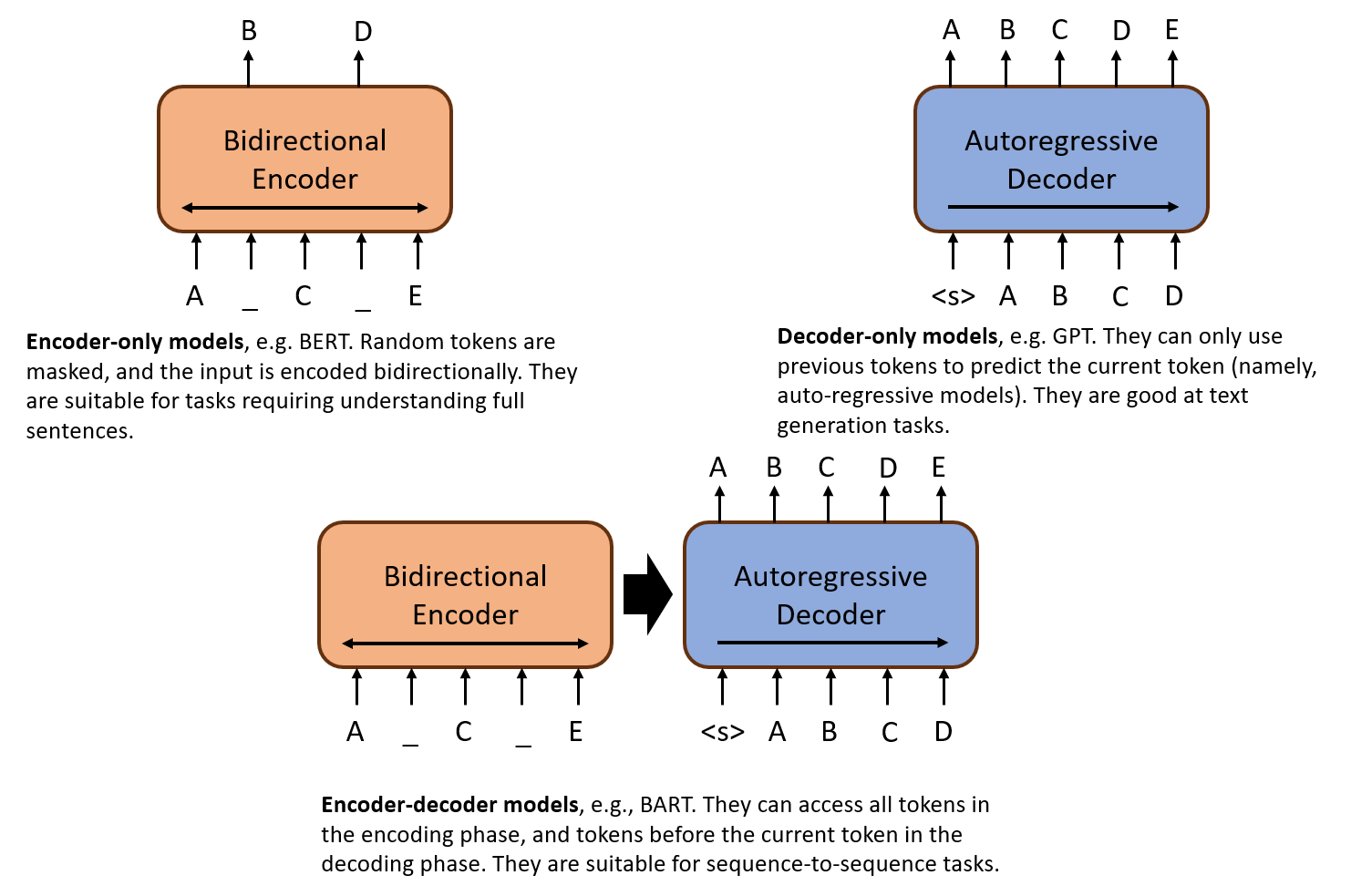}
\caption{Illustration of different transformer models, where BERT is the
encoder-only model, GPT is the decoder-only model, and BART is the
encoder-decoder model.}\label{fig:transformermodel}
\end{figure*}

Based on different purposes, transformers have encoder-only,
decoder-only, and encoder-decoder three variants as shown in
Table~\ref{tab:transformermodel} and Fig.~\ref{fig:transformermodel}.
Encoder-only models can access all positions given an input and utilize
bi-directional contexts to predict tokens.  They
are suitable for tasks requiring understanding full sentences, such as
text classification.  Transformer decoder-only models can only use
previous tokens to predict the current
token (namely, auto-regressive models). They are
good at text generation tasks such as story generation. Transformer
encoder-decoder models can access all tokens in the
encoding phase, and tokens before the current
token in the decoding phase.  They are suitable for
sequence-to-sequence tasks such as translation and summarization. 

\begin{table}[ht]
\small
\begin{center}
\begin{tabularx}{\textwidth}{c|X}
    \hline
    \multirow{3}{*}{\makecell{Encoder-only models \\ (Bidirectional)}} & BERT~\cite{devlin2018bert}\\
    & RoBERTa~\cite{liu2019roberta}\\
    & ELECTRA~\cite{clark2020electra}\\
    \hline
    \multirow{3}{*}{\makecell{Decoder-only models \\ (Unidirectional)}} & PaLM~\cite{chowdhery2022palm}\\
    & GPT-1, 2 and 3~\cite{radford2018improving, radford2019language, brown2020language}\\
    & Transformer XL~\cite{dai2019transformer}\\
    \hline
    \multirow{2}{*}{\makecell{Encoder-Decoder models \\ (Sequence to sequence)}} & BART~\cite{lewis2020bart}\\
    & T5~\cite{raffel2020exploring}\\
    \hline
\end{tabularx}
\end{center}
\caption{Transformer-based PLMs.}\label{tab:transformermodel}
\end{table}

\section{Pre-trained Language Models}\label{LMsec:PLM}

Pre-trained language models (PLMs) are dominating in the NLP field
nowadays.  With the development of deep learning, the training and usage
of PLMs have changed a lot as compared with conventional statistical
LMs.  Before being applied to real-world tasks, PLMs are first
pre-trained on massive collections of corpora so that they learn
universal representations that carry both syntactic and semantic
knowledge. After pre-training, PLMs are fine-tuned for downstream tasks so
that the acquired knowledge can be transferred to different tasks.  In
the following, we first explain the pre-training objectives in Sec.
\ref{LMsubsec:pretraining} and then talk about how to adapt PLMs to
various tasks of interest through fine-tuning in Sec.
\ref{LMsubsec:finetune}.  It is also worthwhile to point out several
good survey papers on PLMs, e.g.,~\cite{qiu2020pre, han2021pre,
liu2023pre}. 

\subsection{Pre-training}\label{LMsubsec:pretraining}

The most commonly used pre-training task is ``missing
token prediction''.  There are other pre-training
tasks for different purposes, e.g., next-sentence prediction, which
allows an LM to learn sentence relationships. 

{\bf Token Prediction.} Auto-regressive language
LMs~\cite{radford2018improving, radford2019language, brown2020language}
are trained to predict the next token using
previous tokens. While bidirectional
LMs~\cite{devlin2018bert, lan2019albert, liu2019roberta} mask a subset
of tokens in a sample and learn to predict such
masked tokens using the rest of the context.  For
the latter, the most popular objective is the masked language model
(MLM) objective as proposed in BERT~\cite{devlin2018bert}. The MLM
objective is the cross-entropy loss in predicting masked tokens. It
randomly masks out 15\% of the input tokens and then predicts the masked
tokens. The number of masked tokens is set to 15\% based on experimental
verification.  If the masking rate is too small, the model only learns
from a limited number of masked tokens. On the other hand, if it is too
large, there is not enough context to do reasonable predictions and
models cannot learn well. 

{\bf Other Pre-training Tasks.} There are other pre-training tasks to
make LMs learn better linguistic knowledge such as sentence
relationships. For example, next sentence prediction is used as the
pre-training task in BERT~\cite{devlin2018bert}. Next sentence
prediction is formalized as a binary prediction task that decides
whether two sentences are two consecutive sentences or not. In this way,
a PLM can be used in downstream tasks that require the understanding of
the relationship between two sentences, such as Question Answering (QA)
and Natural Language Inference (NLI).  Other pre-training objectives are
adopted by BART~\cite{lewis2020bart}. They include token deletion, text
infilling, sentence permutation, and document rotation to corrupt the
original sequence for reconstruction.  Shuffled tokens are used in
T5~\cite{raffel2020exploring} to increase the robustness of the learned
representation. 

\subsection{Fine-Tuning, Adapter Tuning and Prompt Tuning}\label{LMsubsec:finetune}

PLMs learn non-task-specific language knowledge in the pre-training
stage. Fine-tuning performs task-specific adaptations of the model so
that they can be applied to different downstream tasks.  The model
parameters are updated in the fine-tuning stage. One approach is to
design task-specific heads based on different label spaces and losses in
different downstream tasks, then update the entire model and
task-specific heads.  For instance, GPT
\cite{radford2018improving} and BERT \cite{devlin2018bert} added an
extra linear output layer as task-specific heads in their original
papers, and fine-tuned the entire set of parameters in the PLMs and the
heads for various downstream tasks, such as natural language inference,
question answering, semantic similarity, and text classification. To
make the fine-tuning mechanism more parameter efficient, one can choose
to only update certain layers of an LM and the task-specific heads.

Adapter tuning \cite{houlsby2019parameter,
pfeiffer2021adapterfusion,hulora} is proposed to make fine-tuning even
more parameter efficient compared with updating the last layers of a PLM
only. It injects additional compact layers, calls adapters, into the
original PLMs. Then, the new adapter layers are updated, while the
parameters of the original PLMs are frozen during adapter tuning. In
this way, the parameters of the original PLMs can be shared by different
downstream tasks.

PLMs are pre-trained by one or several pre-training objectives and,
then, applied to different downstream tasks. The gap between
pre-training tasks and downstream task-specific fine-tuning can be
substantial.  Prompt-tuning \cite{liu2023pre} is used to discover the
potential of PLMs by mimicking the pre-training objectives in the
fine-tuning or inference stage.  As PLMs get more powerful, they can
handle various downstream tasks by seeing a few examples without any
gradient updates or fine-tuning. This is achieved by prompt-based
fine-tuning (or prompt-tuning in short). 

The prompt can be divided into discrete prompts
(also called hard prompts) and continuous prompts (also called soft
prompts). A discrete prompt is a natural text template that could be
manually designed by humans \cite{brown2020language,
schick2021exploiting, schick2021few} or automatic methods
\cite{gao2021making, zhou2022large, pryzant2023automatic}. On the
contrary, continuous prompts \cite{li2021prefix, zhong2021factual,
lester2021power, qin2021learning} are continuous vectors in the
embedding space that do not correspond to real text. It sacrifices
interpretability but relaxes the discrete prompt constraint in 
that prompts should be real texts.

\begin{figure*}[t] 
\centering
\includegraphics[width=1\linewidth]{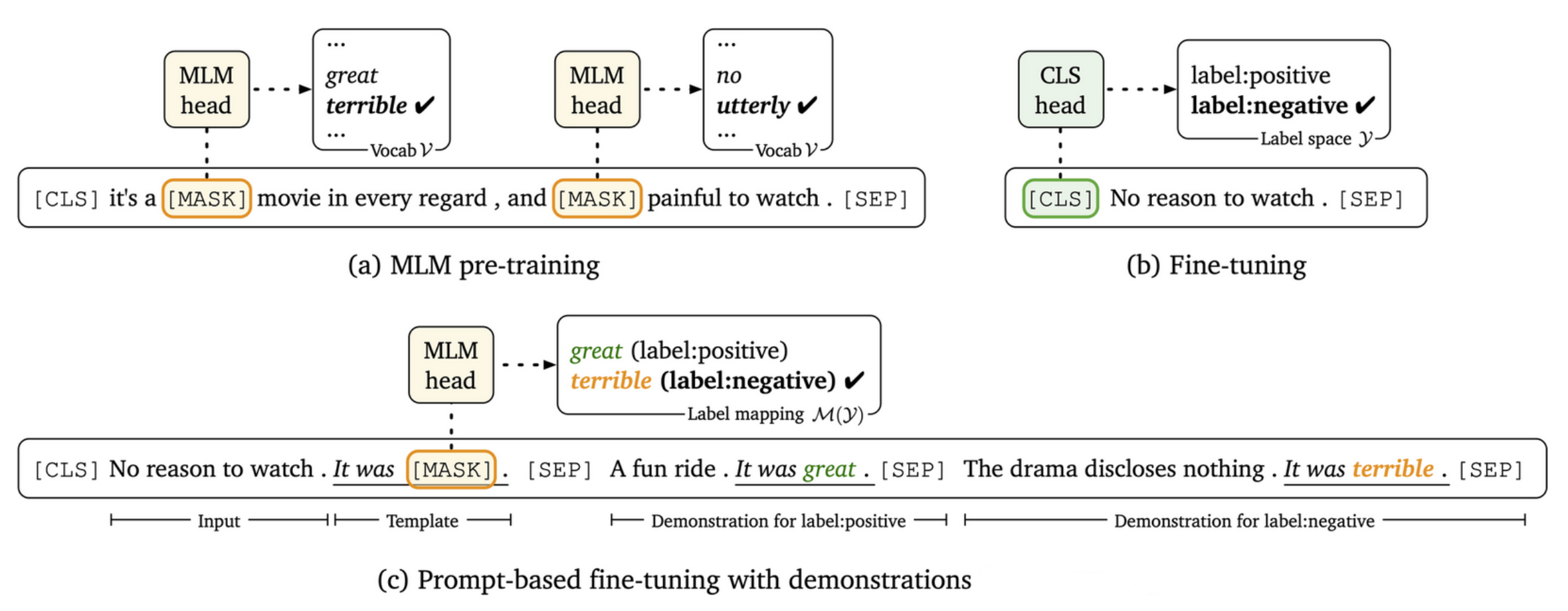} 
\caption{An illustration of (a) LM pre-training, (b)
standard fine-tuning, and (c) discrete prompt-based fine-tuning (or prompt-tuning) 
\cite{gao2021making}.}\label{fig:prompt-tuning}
\end{figure*}

Fig.~\ref{fig:prompt-tuning} shows an example of the pre-training task,
fine-tuning and discrete prompt-tuning of MLMs.  In the pre-training,
MLMs are trained to predict masked tokens.
Assuming that the downstream task is the sentiment analysis of the movie
review. In standard fine-tuning, we train a new head on the top of a PLM
and predict the sentiment labels.  The original input appended with a
designed prompt, say, `It was', is sent to the PLM. The PLM has to
assign probabilities to designed answers, which can be `great' or
`terrible'. If the probability of `great' is higher, then the label of
the input will be positive and vice versa. In this way, prompt-tuning
converts a distinct downstream task to the token
prediction task to narrow the gap between the pre-training and
fine-tuning stages. 

\section{Model Evaluation}\label{LMsec:Evaluation}

There are two LM evaluation types: intrinsic evaluation and extrinsic
evaluation. The intrinsic evaluation examines the internal properties of
an LM while the extrinsic evaluation studies its performance in
downstream tasks. 

\subsection{Intrinsic Evaluation}\label{LMsubsec:Intrinsic_Evaluation}

{\bf Auto-regressive LM.} LMs estimate the probability of text
sequences. A good LM assigns higher probabilities to natural text
sequences and lower ones to unreal or random text sequences. The
perplexity is a common evaluation metric for this purpose. Given a testing
text sequence, the perplexity, denoted by $PPL$, is defined as the
inverse probability of the sequence normalized by the number of 
tokens. Mathematically, we have
\begin{equation} \label{eq:perplexity1}
PPL(S) = \sqrt[N]{\frac{1}{(P(u_1u_2...u_N)}},
\end{equation}
where $S = u_1u_2 ...u_N$ is a testing text sequence.  The perplexity can
be rewritten in form of 
\begin{equation} \label{eq:perplexity2}
PPL(S) = \sqrt[N]{\prod_{i=1}^N\frac{1}{P(u_i|u_1...u_{i-1})}}.
\end{equation}
A good LM should maximize the text set probability. It is equivalent to
minimizing the perplexity. The lower the perplexity, the better the LM. 

{\bf Bidirectional Language Model.} To calculate the inverse probability
in Eq. (\ref{eq:perplexity1}), the auto-regressive LMs can use a
sequence of conditional probabilities.  However, this approach does not
work for bidirectional LMs (or masked LMs). Several intrinsic evaluation
metrics have been proposed for bidirectional LMs.  The
pseudo-log-likelihood score (PLL)~\cite{wang2019bert} is defined as
\begin{equation} \label{eq:pll}
PLL(S) = \sum_{i=1}^{|S|}\log{P(u_i|S_{\backslash i})},
\end{equation}
where $\log{P}(u_i|S_{\backslash i})$ is the conditional probability of
token $u_i$ in sentence $S$ with all remaining tokens.  Instead of
maximizing the joint probability of the entire text sequence, a good
bidirectional LM should maximize the probability of each token in the
text sequence given other tokens.  Based on PLLs, the pseudo-Perplexity
(PPPL) for corpora $C$ is defined as~\cite{salazar2020masked}
\begin{equation} \label{eq:pppl}
PPPL(C) = exp(-\frac{1}{N}\sum_{S\in C}^{}PLL(S)).
\end{equation}
Both PLL and PPPL provide effective means to measure the naturalness of
sentences for a bidirectional LM. For example, it was shown
in~\cite{salazar2020masked} that PLL and PPPL correlate well with the
performance of an LM on downstream tasks, such as automatic speech 
recognition and machine translation. 

\subsection{Extrinsic Evaluation}\label{LMsubsec:Extrinsic_Evaluation}

Any downstream task of LMs can be used for extrinsic evaluation.  There
are several common downstream tasks selected as extrinsic evaluation
benchmarks. Two popular ones are GLUE (General Language Understanding
Evaluation) \cite{wang2018glue} and SuperGLUE \cite{wang2019superglue}.
GLU is an evaluation benchmark for natural language understanding. It
contains single-sentence tasks, similarity and paraphrase tasks, and
inference tasks.  SuperGLUE is an enhanced version of GLUE. It includes
a new set of more challenging language understanding tasks, more diverse
task formats, improved resources, and a public leaderboard. 

\subsection{Relation between Intrinsic and Extrinsic Evaluations}
\label{LMsec: relation_in_and_ex}

If an LM achieves a lower perplexity, does that mean it can also perform
well on downstream tasks? In other words, is there any correlation
between pre-training tasks (based on token
prediction) and the downstream tasks?  There are many empirical studies
on this question but few theoretical studies. 

{\bf Empirical Studies.} Researchers design experiments to understand
what kind of knowledge is learned by an LM from the pre-training tasks.
Examples include ~\cite{tenney2019you, giulianelli2018under,
tenney2019bert, kim2020pre, hewitt2019structural, rogers2020primer}.
They use part-of-speech tagging, constituent labeling, and dependency
labeling to measure the degree of syntactic knowledge learning, and
named entity labeling, semantic role labeling, and semantic proto-role
for testing semantic knowledge.  Empirical studies show that
pre-training tasks help LMs learn the linguistic knowledge such as the
grammar~\cite{kim2020pre} and the semantic role~\cite{rogers2020primer}.
However, these experimental results can only be used as evidence
supporting that the token prediction tasks benefit
downstream tasks. They cannot explain the underlying mechanism. 

{\bf Theoretical Studies.} Some researchers attempt to build the connection
between LM's perplexities and its performance on downstream tasks
mathematically. The text classification tasks were studied
in~\cite{saunshi2020mathematical}.  They first hypothesized and verified
that text classification tasks can be reformulated as sentence
completion tasks. Since the LM pre-training task is essentially a
sentence completion task, it does help the text classification downstream
task.  Then, they quantified the connection mathematically and showed
that the features from LMs that achieve $\epsilon$-optimal in
log-perplexity can linearly solve text classification tasks with
$O(\sqrt{\epsilon})$ error.  An underlying generative model was utilized
in \cite{wei2021pretrained} to show the relationship between the
pre-training tasks and the downstream tasks.  Current theoretical
studies are limited in the sense that only a specific downstream task
(say, the text classification task) is considered and the proof holds
under certain conditions. 

\subsection{Beyond Single Metric for LM Evaluation}
\label{LMsec: other_eva}

Except for the evaluation of LM's performance on
standard evaluation test sets, the LM performance on other aspects is
also important in real-world applications, such as efficiency
\cite{strubell2019energy, thompson2020computational, bender2021dangers,
wu2022sustainable}, bias \cite{may2019measuring, blodgett2020language,
nadeem2021stereoset, abid2021persistent}, robustness
\cite{rychalska2019models, omar2022robust, jin2020bert, zang2020word,
nie2020adversarial, goel2021robustness, wang2021textflint},
explainability \cite{zini2022explainability}, and logical consistency
\cite{ribeiro2019red}. In this section, we discuss evaluations
on efficiency, bias, and robustness to provide a holistic review of
evaluation aspects.

\textbf{Efficiency} of LMs can be evaluated in
several aspects, such as inference time, computational complexity,
energy consumption, model size, and training data size.  Some work
\cite{strubell2019energy, wu2022sustainable, thompson2020computational,
wang2023overview} calculated the computational complexity, approximate
financial, and environmental costs of training PLMs. They also suggested
practical steps to reduce expenses in NLP research and applications.
Discussion on the model size of recently developed PLMs was given in
\cite{bender2021dangers}. In Sec. \ref{LMsec:Efficient_LMs} of this
paper, we also discussed several methods to achieve efficient LMs.
Table~\ref{tab: model size} shows the number of parameters, training
data, cost, and time of recently developed LMs.

\textbf{Bias} in NLP refers to systematic prejudices
of models resulting from erroneous assumptions, such as racism, sexism,
and ableism. Bias is reflected in PLMs since they are trained on a large
volume of real word data. Several studies have examined bias in PLMs.
The Sentence Encoder Association Test (SEAT) was proposed in
\cite{may2019measuring} to investigate bias in BERT
\cite{devlin2018bert}.  A dataset was created in
\cite{nadeem2021stereoset} to measure bias against gender, profession,
race, and religion across multiple PLMs, including BERT
\cite{devlin2018bert}, RoBERTa \cite{liu2019roberta}, XLNet
\cite{yang2019xlnet} and GPT-2 \cite{radford2019language}.  It was
demonstrated in \cite{abid2021persistent} that GPT-3
\cite{brown2020language} consistently exhibits a significant anti-Muslim
bias in various tasks. The work in \cite{blodgett2020language} surveyed
146 papers on bias in NLP and made recommendations for analyzing bias in
NLP systems.

\textbf{Robustness} of LMs refers to their capacity
to perform effectively and consistently when confronted with input
variations (e.g., typos and misspellings) that should not affect the
system's output. In other words, a robust LM should not be easily fooled
by adversarial text. Recent studies\cite{jin2020bert, zang2020word,
nie2020adversarial} created a set of character or word level
perturbations to simulate various types of noise that LMs may encounter
in real-world scenarios. They examined robustness of recently developed
PLMs, including BERT, RoBERTa and XLNets. The results suggest that input
perturbations, even minor alterations, can harm the performance of these
LMs. In addition, Robustness Gym \cite{goel2021robustness}, WildNLP
\cite{rychalska2019models}, and TextFlint \cite{wang2021textflint} are
tools designed for robustness evaluation.

\section{Language Models in Text Generation}\label{LMsec:Application}

One of the most important applications of LMs is text generation, which
aims to generate sequences of words based on the input data. There are
many text generation tasks because of different purposes and inputs. For
example, the automatic speech recognition (ASR) task demands that the
input be a speech sequence while the output be the corresponding text
sequence. The machine translation task generates the translated text
sequence based on the input text sequence and the target language. Story
Generation is a topic-to-text generation task.  In this section, we
introduce common techniques used in text generation and then explain how
LMs can be applied in each of the representative tasks. 

\subsection{Decoding Methods}\label{LMsubsec:Decoding}

\begin{figure*}[ht] 
\centering
\includegraphics[width=1\linewidth]{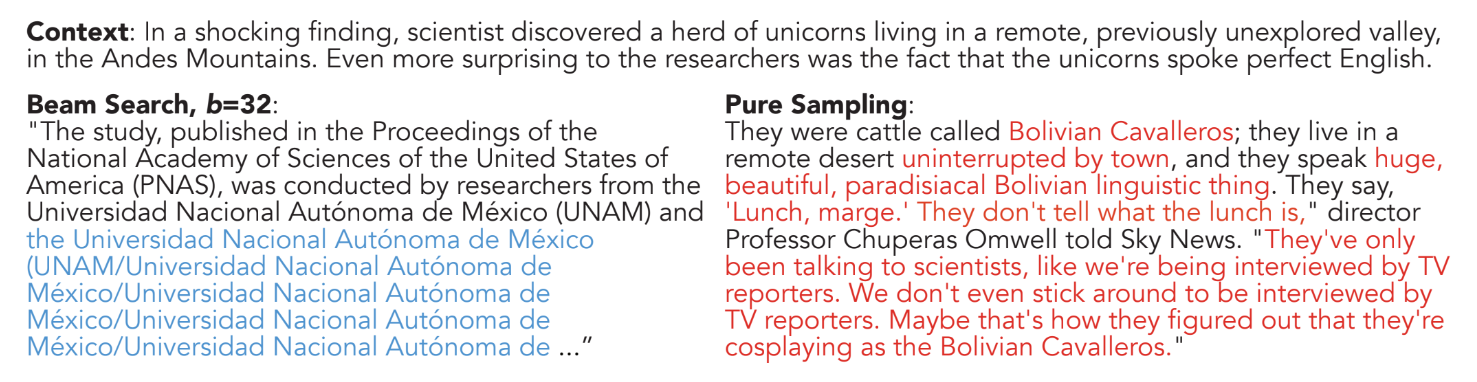} 
\caption{Comparison of texts generated by the powerful GPT-2 large
language model (LLM) using Beam search (left) and pure sampling decoding
(right). Beam search yields degenerate repetition (in blue) while pure
sampling results in incoherent gibberish (in red)
\cite{holtzman2019curious}.} \label{fig:nlg}
\end{figure*}

Decoding decides the next output linguistic unit to generate text. A good decoding method should generate coherent continuation given a
context. As LMs get more sophisticated, decoding methods have played an
increasingly important role. As shown in Fig.~\ref{fig:nlg}, deficient
decoding methods lead to bad generated texts even with a powerful LM.
There are two main decoding methods for text generation. 

{\bf Maximization-based decoding.} This is the most commonly used
decoding objective. Assuming that the model assigns a higher probability
to a higher quality text which is closer to the ground truth written by
humans, the maximization-based decoding strategy searches for tokens
with the highest probability as the generated text. Greedy
search~\cite{zhao2017learning, xu2017neural} chooses the token with the
highest probability as the next token in a greedy manner. Beam
search~\cite{li2016simple, vijayakumar2018diverse,
kulikov2019importance} keeps a certain number of most likely tokens at
each time step and selects the generated token sequences with the
overall highest probability eventually. It avoids missing reasonable
tokens that do not have the highest probability. Trainable decoding
algorithms have been proposed recently. Trainable greedy
decoding~\cite{gu2017trainable} is a neural-based solution that works as
part of a neural machine translation decoder. It utilizes reinforcement
learning to find a translation that maximizes a decoding objective. 

{\bf Sampling-based decoding}. It chooses the next token from a set of
sampled tokens. Because maximization-based decoding depends highly on
the underlying model probabilities and suffers from producing degenerate
repetition, sampling-based decoding increases the diversity of generated
texts by random sampling. However, the simple pure sampling may choose a
token with low probability (from an unreliable tail distribution) as the
next generated token. As a result, the generated text could be unrelated
to the prefix, leading to incoherent gibberish. Top-k sampling
\cite{fan2018hierarchical} and Nucleus sampling
\cite{holtzman2019curious} have recently been proposed to address this
problem. Both Top-k sampling and Nucleus sampling
sample from truncated LM distributions (i.e., sampling from the most
probable tokens). Diverse Beam search~\cite{li2016simple} is a trainable
sampling-based (stochastic) decoding algorithm based on the Beam search.
It uses reinforcement learning to determine the beam diversity
parameters for different inputs or tasks. 

\subsection{Dialogue Systems}\label{LMsubsec:Dialogue}

A dialogue system aims at simulating human responses when conversing
with human users. Recent dialogue systems such as
ChatGPT~\footnote{\url{https://openai.com/blog/chatgpt/}} and LaMDA
~\cite{thoppilan2022lamda} have attracted a lot of attention in the
generative AI field because of their superior performance as interactive
chatbot systems.  Dialogue systems can be categorized into task-oriented
systems and open-domain systems. The former is designed for specific
tasks such as customer service for online shopping. The latter is also
known as chatbots ~\cite{ni2022recent}.  Most modern dialogue systems
are fine-tuned versions of generative LMs.  Taking ChatGPT as an
example, ChatGPT is built based on a generative LM, GPT-3
~\cite{brown2020language} with 175 billion
parameters. It is further fine-tuned by supervised learning and
reinforcement learning on labeled data. 

LMs play an important role in dialogue systems, especially in their
natural language understanding (NLU) and natural language generation
(NLG) components \cite{wang2022focused,wang-etal-2022-analyzing}. NLU is
responsible for understanding and recognizing users' intent. Nowadays,
for encoder-decoder PLMs', the encoders provide
informative representations for NLU, while the associated decoders are
responsible for generating an appropriate response. The latter involves
constructing the response text, selecting appropriate words, and
determining the correct phrasing and tone.  The effectiveness of
representations of PLMs was examined in~\cite{wu-xiong-2020-probing} for
dialogue tasks. The evaluation PLM targets included BERT
\cite{devlin2018bert} and GPT-2 \cite{radford2019language}. The few-shot
capability of PLMs in dialogue tasks such as NLU and NLG was evaluated
in ~\cite{madotto2020language}.  Overall, LMs in dialogue systems play a
key role in understanding users' input and generating appropriate and
natural responses. 

\subsection{Automatic Speech Recognition}\label{LMsubsec:ASR}

Automatic speech recognition (ASR) is a speech-to-text generation task
that aims to transform raw audio input into the corresponding text
sequence. The LM plays an essential role in an ASR system. First, it
helps solve acoustically ambiguous utterances. Second, it can lower the
computational cost by constraining the search space in a set of words of
higher probability. Conventional ASR systems
contain two independent models, an acoustic model and a language model,
which are related by
\begin{equation} \label{eq:ASR}
P(word|sound) \propto P(sound|word)P(word).
\end{equation}
The acoustic model is conditioned on phones $P(sound|word)$ while the LM
gives the word distribution denoted by $P(word)$. LMs help search the
word hypotheses during recognition. Different types of LMs have been
explored in ASR, such as N-gram~\cite{jelinek1990self, siu2000variable},
FFNN~\cite{arisoy2013converting}, RNN~\cite{arisoy2015bidirectional,
huang2014cache} and Transformer~\cite{shin2019effective}

With the development of deep learning techniques,
end-to-end (E2E) ASR systems have emerged as the dominant approach in
this field nowadays. E2E ASR systems do not train the acoustic model and
the language model independently but use a single-network architecture.
For example, the Listen, Attend, and Spell (LAS) model
\cite{chan2016listen} contains an encoder, a decoder, and an attention
network, which are trained jointly to predict the output text sequence.
The LM component in the E2E ASR system is implicitly learned from the
transcribed speech data. To address the challenge of limited transcribed
speech data for LM's training, one solution is to integrate external
language models trained on extensive text corpora using LM integration
\cite{kannan2018analysis, toshniwal2018comparison}. Shallow fusion
\cite{gulcehre2015using, chorowski2017towards, mcdermott2019density}
considers log-linear interpolation between the scores from an E2E ASR
model and an external LM at the decoding stage. Deep fusion
\cite{gulcehre2015using} integrates an external LM and the E2E ASR model
by fusing their hidden states. Unlike shallow fusion and deep fusion,
where the E2E ASR model and the external LM are separately trained, cold
fusion \cite{sriram2018cold} and component fusion
\cite{shan2019component} train the E2E ASR model and the external LM
jointly.

\subsection{Machine Translation}\label{LMsubsec:ML}

Machine translation is a text-to-text generation task where the text in
the source language is translated into that of the target language.  LMs
adopted by machine translation are conditioned on the source sentence
and the previous partial translation. The E2E
machine translation models become prevailing nowadays. The language
model is implicitly learned through E2E training. Recently,
transformer-based models achieve great success in machine
translation~\cite{vaswani2017attention, wang2019learning}.  Similar to
ASR advancements, an external LM trained by extensive monolingual
corpora can be incorporated into an E2E machine translation model
through LM integration techniques~\cite{gulcehre2015using}. Furthermore,
many PLMs have shown their few-shot or zero-shot ability on machine
translation \cite{brown2020language, chowdhery2022palm} although they
have never been explicitly trained on translation parallel data between
the source and the target languages.

\subsection{Detection of Generated texts}\label{LMsubsec:detection}

As the performance of LMs gets closer to or even outperforms humans, the
misuse of LMs, such as fake news and fake product reviews generation,
has become a serious problem. The ability to detect machine-generated
texts is important.  There are two types of detection problems: 1) human
written vs. machine generated, and 2) inveracious vs. veracious.  Most
datasets, e.g., \cite{weiss2019deepfake, uchendu2020authorship,
fagni2021tweepfake}, are collected for the first type. Problems of the
second type are much harder than those of the first type
~\cite{thorne2018automated} since one needs to connect generated text to
the fact, which requires a high-level knowledge reasoning capability.

Two common approaches to detecting machine-generated text are reviewed
below. One is to exploit the probability distribution of
LMs~\cite{ippolito2020automatic, gehrmann2019gltr}. If the probability
distribution of a text sequence is closer to that of human-written texts
as compared with known machine-generated texts, the text sequence is
classified as human-written. The other is to train classifiers with
supervised learning~\cite{zellers2019defending, uchendu2020authorship}. 
It converts the distribution to a supervised binary classification task. 
For more details on the detection of machine-generated texts, readers 
are referred to two survey papers~\cite{jawahar2020automatic, stiff2022detecting}. 

\section{Efficient Models} \label{LMsec:Efficient_LMs}

As recent PLMs get more powerful, their model size, training cost, and
demand for training data increase tremendously.  They need high
computational resources and energy consumption, limiting their real-world applications.  Table~\ref{tab: model size} shows the model size,
training data, cost, and time of recently developed LMs.  This issue is
a concern to many people and the construction of efficient LMs has
received attention.

\begin{table}[h!]
\centering
\begin{adjustbox}{width=1.0\textwidth,center}
\begin{tabular}{|c|c|c|c|c|c|}  \hline
 Model & Year & Number of Parameters & Training data & Training cost & Training time \\ \hline
 BERT-Large & 2018 & 340M & 3.3B words & \$7,000~\tablefootnote{\label{note1} \url{https://syncedreview.com/2019/06/27/the-staggering-cost-of-training-sota-ai-models/}}  & \begin{tabular}{@{}c@{}} 64 TPU chips \\ 4 days \end{tabular} \\ \hline
 XLNet-Lagre & 2019 & 340M & 32.9B tokens & \$245,000~\footref{note1} & \begin{tabular}{@{}c@{}} 512 TPU v3 chips \\ 5.5 days \end{tabular}\\  \hline
 GPT-2 & 2019 & 1.5B & 8 million web pages & \$12,902–\$43,008 \cite{strubell2019energy} & \begin{tabular}{@{}c@{}} 32 TPU v3 chip \\ 168
hours \end{tabular} \\  \hline
 Megatron-LM & 2019 & 8.3B & 174 GB of text data &  & \begin{tabular}{@{}c@{}} 512 GPUs \\ 2 days per epoch \end{tabular} \\ \hline
 T5 & 2019 & 11B & 745GB of text data & Over \$1.3 million \cite{sharir2020cost} & \\ \hline
 Turing-NLG & 2020 & 17B &  &  & \\ \hline
 GPT-3 & 2020 & 175B & 570GB of text data & Over \$4.6 million~\tablefootnote{\url{https://lambdalabs.com/blog/demystifying-gpt-3}}  & \begin{tabular}{@{}c@{}} 1024 A100 GPUs \\ 34 days \cite{narayanan2021efficient} \end{tabular}\\ \hline
\begin{tabular}{@{}c@{}} 
 Megatron-Turing \\ NLG \end{tabular} & 2022 & 530B & 270B tokens &  & \begin{tabular}{@{}c@{}} 2K A100 GPUs \\ 3 months~\tablefootnote{\url{https://www.deepspeed.ai/}} \end{tabular}\\  \hline
\end{tabular}
\end{adjustbox}
\caption{Table of the number of parameters, training data, cost, and time of
several large LMs, where blank cells indicate that the data are not
available. The sources are cited if the data are not obtained from the original work}\label{tab: model size}
\end{table}

\subsection{Data Usage}

\begin{figure}[ht]
\centering
\includegraphics[width=0.65\linewidth]{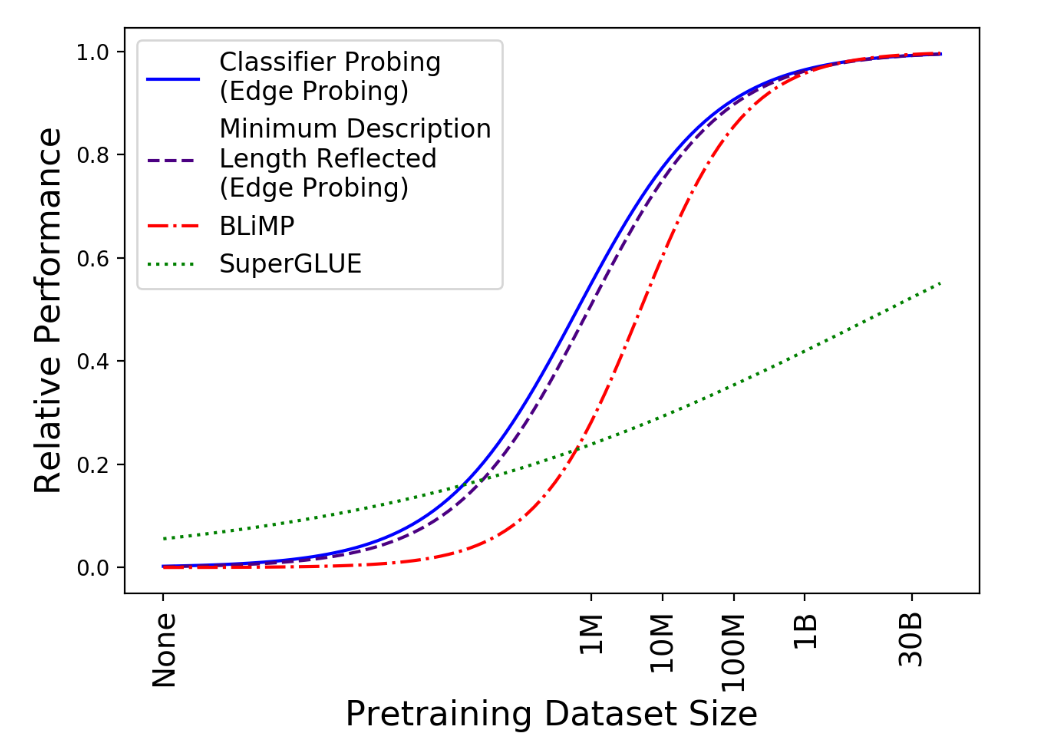}
\caption{The performance curves as functions of the pre-training dataset size,
where the classifier probing measures the quality of the syntactic and semantic
features, the minimum description length probing quantifies the accessibility
of these features, the BLiMP curve measures the model's knowledge of various
syntactic phenomena, and the superGLUE measures the capability of handling 
NLU tasks~\cite{zhang2021you}.}\label{fig:lmtrainingdata}
\end{figure}

{\bf Pre-training Data Size.} A critical question for PLM training is
how much data is needed.  The effect of the pre-training data size on the
RoBERTa model was studied in \cite{zhang2021you}.  The learning curves
of four model performance measures as a function of the pre-training
dataset size are shown in Fig.~\ref{fig:lmtrainingdata}.  When the
data size ranges between 100M and 1B words, three learning curves
gradually level off and it implies that LMs encode most syntactic and
semantic features. However, a much larger quantity of data is needed for
LMs to acquire enough common-sense knowledge and other skills to achieve
better performance on downstream NLU tasks. 

\begin{figure}[ht]
\centering
\includegraphics[width=0.8\linewidth]{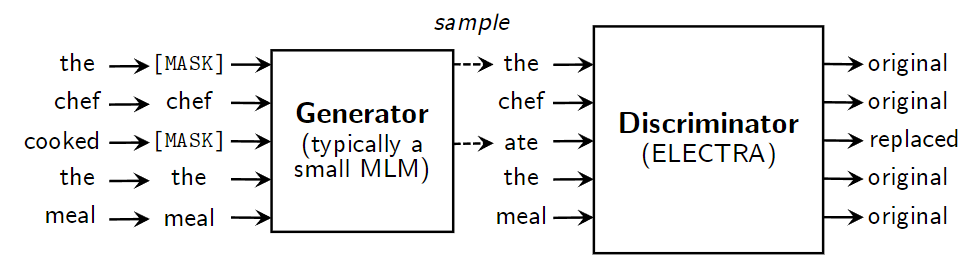}
\caption{The structure of ELECTRA (Efficiently Learning an Encoder that
Classifier Token Replacements Accurately)~\cite{clark2020electra}}
\label{fig:electra}
\end{figure}

{\bf Efficient Pre-Training.} Several methods have been proposed to use
the pre-training data more efficiently.  In the pre-training of masked
LMs, a certain percentage of tokens are masked and need to be inferred
by context. This approach incurs a substantial amount of computational
cost because the network only learns from a certain percentage of tokens
which are masked. To enhance training efficiency, the work
in~\cite{clark2020electra} uses ``replaced token detection" (rather than
``masked token prediction") as the pre-training task.  As shown in
Fig.~\ref{fig:electra}, a generator is trained to perform the masked LM
and predicts the masked tokens. Then, the main model works as a
discriminator, called ELECTRA, which learns to decide the original or
replaced tokens. In this way, pre-training tasks are conducted on all
tokens instead of a small subset of masked tokens.  Learning from all
input positions causes ELECTRA to train much faster than BERT which
adopts masked token prediction. Besides, ELECTRA
achieves higher accuracy on downstream tasks when it is fully trained.
Later, a new pre-training task using an energy-based model, which is
closely related to ELECTRA, is proposed in~\cite{clark2020pre}. 

{\bf Bridging Pre-training and Downstream Tasks.} A typical pre-training
task is token prediction, which often has a large
gap with downstream tasks. To mitigate the gap between pre-training and
downstream tasks, prompt tuning has been studied
in~\cite{radford2019language, gao2021making, schick2021exploiting,
shin2020autoprompt}. As illustrated in Fig.~\ref{fig:prompt-tuning}, the
head is trained to predict the masked tokens in
masked LMs.  For the downstream sentiment analysis task, the head is
trained to predict the positive or the negative label in traditional
fine-tuning.  A template (e.g., `It was') and its expected text
responses (e.g., `great' and `terrible') are used in prompt tuning.  In
this way, pre-training and prompt tuning share the same
``token prediction" objective. 

\subsection{Model Size}

Besides improving training efficiency, efficient LMs focus on the design
of models of smaller sizes. Many methods are investigated to reduce the
model size so that the model can be implemented on mobile or edge
devices with limited computing resources. Model compression is a widely
studied topic.  Compression methods first train a large LM and then
compress it into a target size. Examples include model pruning
\cite{wang2020structured, voita2019analyzing, guo2021parameter},
knowledge distillation~\cite{jiao2020tinybert, sanh2019distilbert,
turc2019well}, low rank matrix approximation~\cite{yang2017breaking,
ma2019tensorized, hu2021lora}, and parameter
sharing~\cite{dehghani2018universal, lan2019albert, dabre2019recurrent,
rothe2020leveraging}. 

\subsection{Inference latency}

Inference efficiency is important to an LM,
particularly in real-time applications. A model of a smaller size
generally has faster inference speed under the same setting. Knowledge
distillation, pruning, and low rank matrix approximation can be employed
to achieve faster inference time while reducing the model size. For
instance, DistilBERT \cite{sanh2019distilbert}, which is a distilled
version of BERT, has demonstrated a 60\% improvement in the inference
speed compared to the original model.  More than 2x speed-up in
inference is achieved in \cite{wang2020structured} by pruning PLMs.

Fast inference speed can also be achieved by fast
decoding methods. Non-autoregressive generation (NAG) models
\cite{gu2018non, su2021non, li2022elmer} predict each token
simultaneously. They have a faster inference speed than autoregressive
models due to parallel computation. On the other hand, the performance
of NAG models is generally worse than autoregressive models since they
do not consider the forward or backward dependency between tokens in the
output text.

\section{Future Research Directions}\label{LMsec:future_directions}

In this section, we describe several promising future research
directions in language modeling. 

\subsection{Integration of LMs and KGs}

Knowledge Graph (KG) provides a structured representation of human
knowledge \cite{wang2022kgboost,ge2022compounde}.  It has been widely
used in many NLP applications, such as question answering
~\cite{huang2019knowledge} and text summarization
~\cite{huang2020knowledge}, because of its capability to represent
relationships between entities. There is a growing interest in
evaluating the knowledge learned in
PLMs ~\cite{petroni2019language, petroni2020how},
where the relationship between different semantic units is captured in
the embedding space and the self-attention layers. Several ideas are
proposed in~\cite{agarwal-etal-2021-knowledge,
zhang-etal-2019-ernie, he-etal-2021-klmo-knowledge, yasunaga2022deep,
zhang2022greaselm, xiong2019pretrained, wang-etal-2021-kepler} to
leverage KGs for LM training.  As a result, the knowledge learned in the
models can be greatly improved. Thus, it is worth careful investigation
of integrating KGs with LMs and understanding how they interact with
each other. 

It appears that KG can serve as an information database to be queried by
LMs. LMs are powerful in natural language understanding and generation
while KGs can organize and store the knowledge information extracted
from the training corpus. In other words, we may decompose knowledge
sources into semantic and syntactic two components, which can be handled
by KGs and LMs, respectively. 

Specifically, most reasoning is handled by KGs so
that predictions are fact-based and explainable. On the other hand, LM
serves as an interface to understand and interpret the language input
and improve fluency, comprehensiveness, conciseness, etc., of the
language output. Similar concepts were proposed in
\cite{yasunaga-etal-2021-qa, hu2022empowering}. In the training phase,
a KG is constructed based on the information extracted from the training
corpus, and an LM can be trained simultaneously. In the inference phase,
an LM can serve as an interface between humans and the knowledge
database represented in the form of KGs.  There are advantages to
assigning semantic and syntactic processing tasks to KGs and LMs,
respectively.  For example, the decoupling facilitates incremental
learning, allows a smaller model size, and improves interpretability.
They will be further elaborated on below. 

\subsection{Incremental Learning}

Incremental learning aims to incorporate new information without
re-training existing models entirely.  The problem of catastrophic
forgetting associated with neural network models was pointed out in
\cite{french1999catastrophic}. That is, the information that has already
been learned by a model can be gradually forgotten when training with
new information. This problem is particularly critical to large LMs
since new information keeps arriving. A solution to catastrophic
forgetting was proposed in \cite{kirkpatrick2017overcoming}.  It
attempts to remember prior important tasks by slowing down learning on
weights that are more relevant to them. However, it is difficult to
define important tasks in LMs. In addition, re-training a large LM with
both old and new data is too expensive.  Lifelong
learning of LMs~\cite{luu-etal-2022-time, lazaridou2021mind,
jin-etal-2022-lifelong} is another solution to accommodate new data to
update the knowledge in LMs. It worths further exploration.

The importance of developing a satisfactory solution to incremental
learning for LMs cannot be over-emphasized. Incremental learning is
challenging for neural networks. Yet, it is easy for KGs to add new data
to (or remove old data from) an existing database by adding or removing
factual triples \cite{wang2021inductive}.  Clearly, the current
information in the KGs will not be overwritten by newly collected data.
The information in the database is updated incrementally. To this end,
the integration of KGs and LMs provides an excellent solution that meets
the need for incremental learning. 

\subsection{Lightweight Models}

As mentioned in Section~\ref{LMsec:Efficient_LMs}, PLMs get more
powerful at the expense of huge computational resources and energy
consumption. The cost issue has to be faced seriously in the development
of large LMs (LLMs). Besides, LLMs are unfriendly to our environment due
to their high carbon footprint.  Green Learning (GL) targets learning
solutions with low carbon footprint. The design of lightweight models of
smaller sizes and lower computational complexity without sacrificing
performance has received more attention in recent years
\cite{schwartz2020green, xu2021survey, kuo2022green, wang2022greenkgc}.  The design of
green LMs is an important topic worth serious investigation. 

Current PLMs are data-driven models that use neural architectures to
learn generic language knowledge from a large amount of data. Efforts have
been made in the development of lightweight LMs.  Model compression is
one of the popular approaches to obtaining a small LM.  Examples include
knowledge distillation or pruning~\cite{li2020train}.  However, this
methodology appears to be a detour since it trains large models and then
shrinks their sizes by compression. Instead, we may incorporate the
linguistic information and the domain knowledge to offer a more direct
way to reduce the model size and the amount of training data. 

\subsection{Universal versus Domain-Specific Models}

A universal LM is developed to handle tasks in the general domain. For
example, ChatGPT is a universal dialogue LM pre-trained on multilingual
and general domain corpora. It can converse on open-domain topics in
multiple languages. In contrast, domain-specific LMs
~\cite{zhang2020bert, gu2021domain, liu2022domain, luo2022biogpt} are
designed to deal with domain-specific tasks, e.g., biomedicine,
economics, musicology, etc. 

A universal LM demands a huge model size, a large number of training
examples, and a tremendous amount of computational resources.  Based on
the scaling law of neural language models~\cite{kaplan2020scaling}, the
inference performance scales as a power-law with the model size, the
dataset size, and the amount of computing used for training.  So far,
the largest PLM contains 540-billion
parameters~\cite{chowdhery2022palm}.  Despite the superior performance
and the flexibility to adapt to multiple tasks, we may wonder whether a
huge universal LM is cost-effective. 

For domain-specific LMs, the amount of training data in need is
significantly lower. It was believed that the general domain PLMs
benefit the training of domain-specific LMs.  However, it is reported in
\cite{gu2021domain} that domain-specific LMs, which were pre-trained
from scratch on in-domain data, can provide a solid foundation for
biomedical NLP. In other words, training a domain-specific LM may not
need a huge amount of general corpora and labeled data.  Domain-specific
LMs to be deployed on task-specific scenarios with less training and
inference efforts expect to receive more attention in the future.

\subsection{Interpretable Models}

Although deep-learning-based LMs are dominating the NLP field, they are
inherently black-box methods without mathematical transparency.  Its
interpretability is of concern. Efforts have been made to explain the
black-box LMs. As mentioned in~\ref{LMsec: relation_in_and_ex},
empirical studies are conducted to understand what PLMs have learned
through experimental design.  However, the progress in this direction
may offer insights but not a satisfactory and clean answer.  Providing
theoretical explanations or establishing explainable LMs is still a
challenging and open issue.  A direction to interpretability is to
design an interpretable learning model from scratch. For example, we may
incorporate KGs with LMs. KG is known to be capable of improving the
interpretability and transparency of the system in many reasoning tasks
such as information retrieval~\cite{dietz2018second} and recommendation
systems~\cite{yang2022knowledge}.  For example,
reasoning paths and data sources can be provided with predictions when
KGs are incorporated for reasoning. It is challenging for LMs to do so.
It is critical to develop an interpretable LM to avoid its hallucination
in natural language generation \cite{ji2023survey}.

\subsection{Machine Generated Text Detection}

The most common application of LMs is text generation. As generative
LM's performance gets closer to or even outperforms humans, these LMs
can be used for malicious purposes such as academic dishonesty,
spamming, targeted bot attacks, and fake news/reviews generation.  How
to determine whether a text is generated by LMs or written by humans is
a big challenge nowadays. A high-performance machine-generated text
classifier can only serve as a reference in real-world applications,
since it has false positives (i.e., human-written texts classified as
machine-generated) and false negatives (i.e., machine-generated texts
classified as human-written). In addition, people may be even more
interested in detecting veracious and unveracious texts. They care more
about whether the text is true or not.  Detecting disinformation could
be more difficult than detecting machine/human-generated text without
assessing the factuality.  Additionally, the factuality may change as
time goes by. It is critical to our society in developing effective
tools to identify malicious usages of generative LMs.  

\section{Conclusion}\label{LMsec:conclusion}

A comprehensive overview of CLMs and their successors, PLMs, was
presented in this paper and a wide range of topics was covered.  First,
different levels of linguistic units were introduced and how linguistic
unit prediction is used to train language models was examined.  Second,
tokenization methods adopted by language models were discussed.  Third,
language model architectures and the training
paradigm of PLMs were reviewed. Fourth, we studied the evaluation and
applications of language models.  Especially, several applications in
the context of text generation were detailed.  Finally, several future
research directions were pointed out. The need for explainable,
reliable, domain-specific, and lightweight language models was
emphasized. 

\bibliographystyle{unsrt}  
\bibliography{mybibfile}  

\end{document}